\documentclass[twoside]{article}

\usepackage{PRIMEarxiv}

\usepackage[utf8]{inputenc} 
\usepackage[T1]{fontenc}    
\usepackage{hyperref}       
\usepackage{url}            
\usepackage{booktabs}       
\usepackage{amsfonts}       
\usepackage{nicefrac}       
\usepackage{microtype}      
\usepackage{lipsum}
\usepackage{fancyhdr}       
\usepackage{graphicx}       

\usepackage{algorithm}
\usepackage{algpseudocode}

\usepackage{amssymb}
\usepackage{amsmath,amsfonts}
\usepackage{mathrsfs}

\usepackage{multirow}
\usepackage{makecell}
\usepackage{booktabs}
\usepackage{algpseudocode}
\usepackage{siunitx}

\usepackage{caption}
\usepackage{subcaption}

\title{The Window Dilemma: \\
Why Concept Drift Detection is Ill-posed - \\
Extended Version}

\author{
  Brandon Gower-Winter, Misja Groen, and Georg Krempl \\
  Utrecht University \\
  8 Hiedelberglaan, Utrecht, 3584 CS, NL \\
  \texttt{b.gower-winter@uu.nl, m.groen@uu.nl, g.m.krempl@uu.nl} \\
}

\begin{document}
\maketitle

\begin{abstract}

Non-stationarity of an underlying data generating process that leads to distributional changes over time is a key characteristic of Data Streams. This phenomenon, commonly referred to as Concept Drift, has been intensively studied, and Concept Drift Detectors have been established as a class of methods for detecting such changes (drifts). For the most part, Drift Detectors compare regions (windows) of the data stream and detect drift if those windows are sufficiently dissimilar.

In this work, we introduce the Window Dilemma, an observation that perceived drift is a product of windowing and not necessarily the underlying data generating process. Additionally, we highlight that drift detection is ill-posed, primarily because verification of drift events are implausible in practice. We demonstrate these contributions first by an illustrative example, followed by empirical comparisons of drift detectors against a variety of alternative adaptation strategies. Our main finding is that traditional batch learning techniques often perform better than their drift-aware counterparts further bringing into question the purpose of detectors in Stream Classification.
\end{abstract}

\section{Introduction}

Many real-world machine learning tasks face the challenge that the underlying data generating process is non-stationary, in particular when data arrives over time in so-called Data Streams \cite{SouzaReisMaletzke2020}. A common strategy to address this problem of Concept Drift (distributional changes over time) is to use only a subset of the data for building (or updating) a model. Typically, this is a window of instances sampled consecutively within an interval. Various Change (or Drift) Detection approaches have been proposed, most of which rely on windows to detect changes between concepts.

The contributions of this paper are threefold:
First, in Section \ref{windowdilemma} we show that there is a dilemma in choosing the correct instance selection procedure. Instance selection innately determines the perception of concepts and drifts, which might not match the actual ones.
Second, in Section \ref{driftdetection} we illustrate why drift detection is ill-posed. It depends itself on the selected sample to be an accurate proxy of the current state (concept) of the underlying data generating process. To support this claim, we integrate several well known limitations of drift detection such as the inability to verify all drift in practice. Third, our empirical evaluation (Section \ref{exp}) of various drift-aware adaptive classifiers and drift detectors against drift-unaware classifiers indicates that the type of classifier  is often more important than drift-awareness.
We conclude by providing directions for future research, in particular by moving towards techniques that allow identifying change patterns that allow specific actions and a more stringent evaluation. 

The source code is available at \url{https://github.com/BrandonGower-Winter/TheWindowDilemma}.

\section{Background and Related Work} \label{relwork}

A data stream may be defined as a mode of access to a potentially infinite sequence of instances generated by some concept and delivered to a learning algorithm by an instance-delivery-process (IDP) \cite{ReadZliobaite2025}. Streaming data exists in many real world settings such as sensor, stock market and financial data \cite{SouzaReisMaletzke2020}.

In traditional batch machine learning, we assume the underlying data generating process $\mathcal{D}$ remains constant throughout the training, testing and deployment of a model \cite{HinderVaquetHammer2024}. That is to say that $n$ i.i.d. samples are taken from a stationary distribution, that is, $X_0, X_1, ..., X_n  \sim\mathcal{D}$.\footnote{In the Supervised Learning setting $X_i$ may be written as $(X_i, Y_i)$.}

On the other hand the data stream in a stream learning task is not time-invariant, so each sample is potentially sampled from a unique distribution $X_i \sim \mathcal{D}_i$. Concept drift, a change in the underlying data generating process through time, occurs when for two timesteps $i,j$, $\mathcal{D}_i \neq \mathcal{D}_j$. Given a stream learning task $\mathcal{D}_i = P(X,Y)_i$, concept drift often occurs in the class conditional probabilities $P(Y|X)$ or the covariates $P(X)$.
Concept drift may be categorized by its observed behavior: If a concept changes suddenly, it is known as abrupt drift. If it changes slowly over time, it is known as incremental or gradual drift. If these changes are repetitive (e.g. public transportation usage patterns over workweeks compared to weekends), the drift is known as reoccurring drift \cite{LuLiuDong2018}.

\subsection{Detection of Drift with Change Detectors}

Naturally, if Concept Drift describes the change in a distribution generating process $\mathcal{D}_t$ over time $t$, then a Change (or Drift) Detector is a tool for its detection. Informally, a drift detector is a binary classifier that observes some number of samples, and predicts if the underlying data distribution that produced said samples changed through time.

The drift detection process typically consists of 4 stages \cite{LuLiuDong2018, HinderVaquetHammer2024}. The first stage is data acquisition which refers to the selecting of instances to pair with a drift detector. The most common solutions are window-based instance selection with the sliding window technique being the most popular \cite{LuLiuDong2018}. Alternative options including forgetting mechanisms \cite{ReadZliobaite2025}.

The second stage is the data modeling stage whereby descriptors are built. This stage is optional and only relevant if some kind of preprocessing is necessary for the drift detector to function. Examples of such techniques include dimensionality reduction and binning of numerical values \cite{HinderVaquetHammer2024}.

In Stage 3, the detector computes the dissimilarity of the descriptors, and Stage 4 calibrates the dissimilarity which identifies drift in the selected instances. We are intentionally vague about the last two stages because they are not the focus of this paper. A typical drift detector will either compute some magnitude of dissimilarity \cite{HoensChawlaPolikar2011, GoldenbergWebb2019} or conduct a statistical test \cite{LuLuLiu2025}. 

Interestingly, few research endeavors focus on Stage 1 (Instance Selection) despite it being one of the most important factors in detecting concept drift \cite{KunchevaZliobaite2009, HinderArteltHammer2020, LukatsZielinskiHahn2025}. In window-based methods, this often comes down to choosing the appropriate window size. 
Small windows offer reactivity, while larger windows offer stability. Furthermore, the speed of the concept drift in a data stream will influence the window size. Fast drift needs to be detected quickly (small window), while slow drift requires larger windows \cite{HinderArteltHammer2020}.

Drift Detection research that consider window sizes are Lazarescu et al. \cite{LazarescuVenkateshBui2004} who keep multiple reference windows of different sizes. The detectors ADWIN \cite{BifetGavalda2007}, and PUDD \cite{LuLuLiu2025} use adaptive windowing techniques to be robust to concept drift at different time scales. It is worth noting though that the vast majority of drift detectors simply have the window size as a tunable parameter. It should be emphasized that any approach that relies on a sample of instances for change detection (or classification) inherently depends on the window in which the corresponding instances were observed.

Drift detectors can be categorized in several ways. First, some drift detectors are designed for Supervised or Unsupervised Learning tasks. DDM \cite{GamaMedasCastillo2004}, ADWIN \cite{BifetGavalda2007}, and PUDD \cite{LuLuLiu2025} are examples of the former, and D3 \cite{GozuaccikBuyukccakirBonab2019}, IBDD \cite{SouzaParmezanChowdhury2021} are examples of the latter. Unsupervised Drift Detectors can be used in both Supervised and Unsupervised Learning tasks with some detectors, such as DAWIDD \cite{HinderArteltHammer2020}, being designed with both settings in mind. 

Both Supervised and Unsupervised techniques are limited in the types of drift they can detect \cite{Zliobaite2010}. Supervised (performance-based) detectors often rely on the presence of labeled data which may not be readily available. 
Some techniques such as STUDD \cite{CerqueiraGomesBifet2023} and PUDD \cite{LuLuLiu2025} aim to work around this limitation, but a performance-based drift detector is only reliable if changes in the performance of the deployed predictor reliably indicate drift \cite{KunchevaZliobaite2009,HinderVaquetBrinkrolf2023, ReadZliobaite2025}. 
Conversely, Unsupervised (distribution-based) detectors can only detect changes in the features $P(X)$ and not in the class-conditional probabilities $P(Y|X)$.  For further reading, we refer the reader to the following survey papers \cite{GamaZliobaiteBifet2014, LuLiuDong2018, BayramAhmedKassler2022, HinderVaquetHammer2024, LukatsZielinskiHahn2025, ReadZliobaite2025}.

\subsection{Adaptive Models}

Drift detectors are often used to create adaptive models \cite{LuLiuDong2018}. The exact method for imbuing a model with adaptivity is both model and detector dependent, but the general process is that when a drift event is detected, the deployed model will reset itself in some capacity. This can be through a complete model reset (start training from scratch) or by intelligently altering some aspects of the model's internal structure as is the case with many of the Hoeffding Tree \cite{HultenSpencerDomingos2001} variants. For example, the Hoeffding Adaptive Tree \cite{BifetGalvalda2009} will use a detector to monitor the performance of different subtrees and replace them if necessary.

There also exist ensemble methods such as Adaptive Random Forests \cite{GomesBifetRead2017} and the drift aware variants of Bagging and Boosting \cite{OzaRussell2001, BifetHolmesPfahringer2009} which will replace poorly performing base learners.
It is also possible to have adaptive models that do not utilize drift detectors \cite{ReadZliobaite2025} such as the Aggregated Mondrian Forest \cite{MourtadaGaiffasScornet2021}. Adaptivity may also be added by employing simple reset or retraining regimes \cite{Bifet2017}.

\section{The Window Dilemma} \label{windowdilemma}
The first stage of Drift Detection is instance selection for which windowing methods are the most popular \cite{LuLiuDong2018, LukatsZielinskiHahn2025}. This Section introduces the Window Dilemma, an observation that illustrates why the act of selecting any finite set of past instances (windowing), at best, provides a surrogate representation of the current concept. We also highlight that, in practice, Concept Drift is usually undetectable from the observation of the data stream alone.

Recall that the generating distribution of a data stream is not time-invariant, so each sample is potentially sampled from a unique distribution $X_i \sim \mathcal{D}_i$. Concept drift occurs when for two timesteps $i,j$, $\mathcal{D}_i \neq \mathcal{D}_j$. 

When attempting to detect if $\mathcal{D}_{i} \neq \mathcal{D}_j$, a time window $W(i, j) = \{X_i, ...,X_j\}$ captures the instances between timesteps $i$ and $j$. Ideally, we want to select some intermediate timestep $i < k < j$ such that $\mathcal{D}_{W(i,k)} \approx \mathcal{D}_i$ and $\mathcal{D}_{W(k,j)} \approx \mathcal{D}_j$ which leads to $\mathcal{S}(W(i,k), W(k,j)) \approx \mathcal{S}(\mathcal{D}_i, \mathcal{D}_j)$ for some dissimilarity metric $\mathcal{S}$.

Herein lies the essence of the Window Dilemma: selecting the right $k$ is unlikely without access to the underlying data generating processes $\mathcal{D}_i$ and $\mathcal{D}_j$.
In practice, a window is often simply subdivided in half or an adaptive windowing technique is used to evaluate the window over multiple partitions \cite{BifetGavalda2007}.

Given this, any conclusions drawn from drift detection or analysis are conditioned on the instances selected and how they are partitioned, not necessarily the underlying data generating process.

\begin{figure}[ht]
    \centering
    \includegraphics[width=\columnwidth]{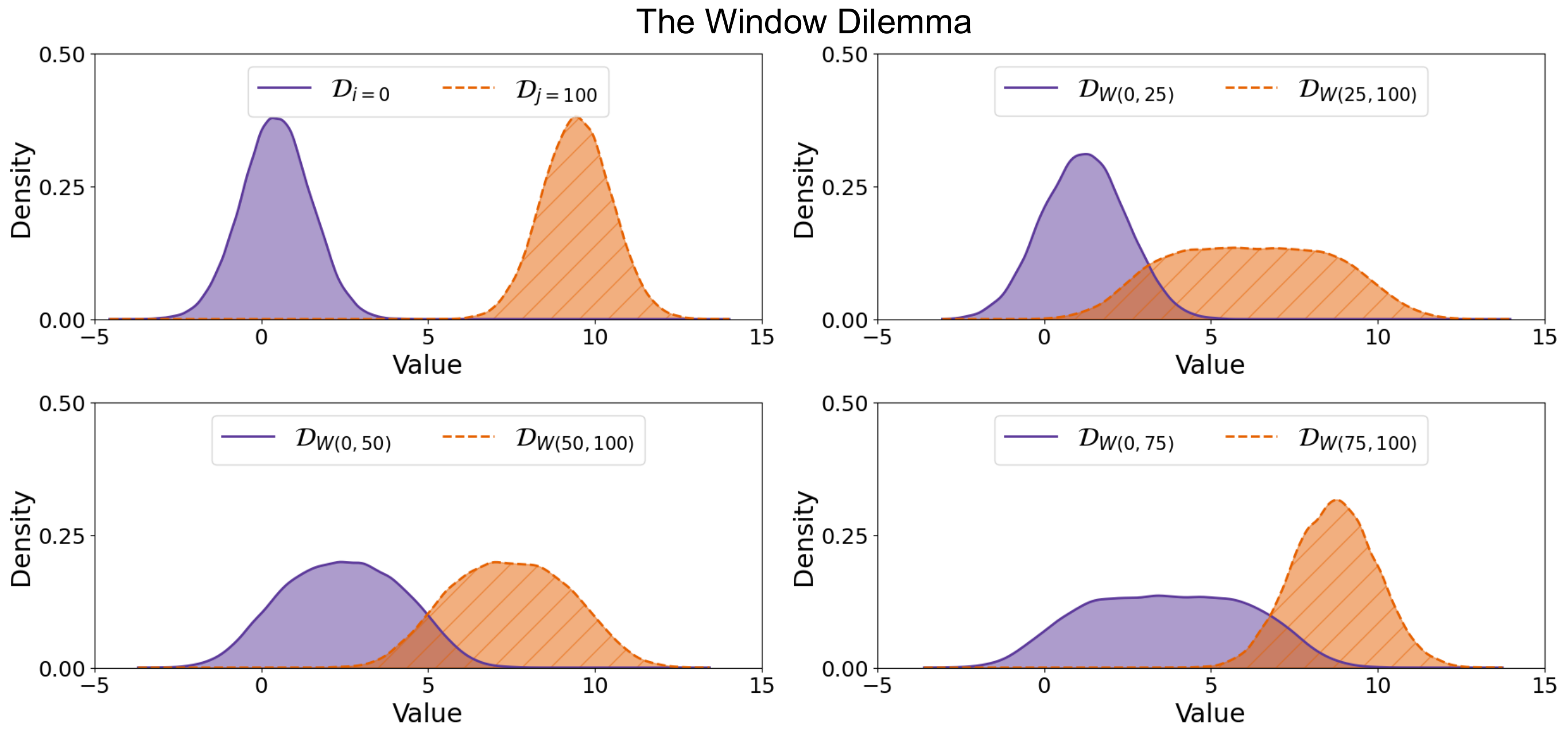}
    \caption{An illustrative example of the Window Dilemma. In the top left we have two distributions $\mathcal{D}_{i}$ and $\mathcal{D}_{j}$ for which we would like to detect Concept Drift. In practice, we have to achieve this by observing samples generated over the timesteps $i=0$ and $j=100$ and keeping them in a Window $W(i,j)$. By choosing some intermediate point $k$, we can construct two sub-windows which we would use to detect drift. The Window Dilemma shows that, depending on our choice of $k$, we will observe varying distributions, none of which are the distributions we are actually trying to compare. In this Incremental Drift setting, choosing a $k$ close to either $i$ or $j$ will approximate the generating distributions of either $\mathcal{D}_{i}$ (top right figure) or $\mathcal{D}_{j}$ (bottom right figure) accurately, at the cost of poorly approximating the other. In the case where $k$ is chosen as the mid-point (bottom left figure), as often done in practice, we see that we get the right shape for each distribution, but the overlap between them isn't representative of the overlap between the actual distributions $\mathcal{D}_{i}$ and $\mathcal{D}_{j}$.}
    \label{fig:Window_Dilemma}
\end{figure}

Consider the following example where we have a data stream containing a single random variable that is sampled from a Gaussian such that $X_i \sim \mathcal{D}_i = \mathcal{N}(\mu_i, \sigma^2)$ where $\sigma = 1$. We induce incremental concept drift by altering $\mu_i$ every $10$ units of time such that $\mu_i = \frac{i}{10}$. Figure \ref{fig:Window_Dilemma} illustrates how various window partitions generate different approximations of distributions $\mathcal{D}_i$ and $\mathcal{D}_j$. When $k \rightarrow j$, the approximated distribution $\mathcal{D}_{W(k, j)} \rightarrow \mathcal{D}_j$. Conversely, when $k \rightarrow i$, the approximated distribution $\mathcal{D}_{W(i, k)} \rightarrow \mathcal{D}_i$. Here we see the Window Dilemma in action, as the choice of window partitioning affects the accuracy of the approximated distributions at timesteps $i$ and $j$. One can construct such scenarios for a myriad of drift types. In the abrupt drift setting, the accuracy at which one approximates distributions $\mathcal{D}_i$ and $\mathcal{D}_j$ is entirely dependent on how precisely the partition point $k$ approximates the moment at which the abrupt drift occurs.

The Window Dilemma is inherent to instance selection over a non-stationary distribution. Thus, it also applies to models that do not use windows explicitly, but implicitly through their instance selection mechanism. 
We are intentionally vague about the number of samples considered in these streams, because it affects how accurately the generating distributions can be represented. We are also not commenting on whether a drift detector would detect drift in these cases. Even the approximated distributions in Figure \ref{fig:Window_Dilemma} are clearly different and it is possible that a drift detector would pick up on this. Whether that is useful is a different matter which we address in the next Section.

\section{Concept Drift Detection is Ill-Posed} \label{driftdetection}

This Section aims to demonstrate why concept drift detection is ill-posed. Formally, a Concept Drift Detector $\mathcal{M}$ may be defined as mapping a window of instances $W(i,j)$ over timesteps $i$ and $j$ to an outcome: $\mathcal{M}: W(i,j) \mapsto \{0, 1\}$. In essence, a drift detector is seen as a binary classification task where the goal is to correctly identify whether concept drift has occurred at a given timestep. This claim is further supported by the use of Accuracy, AUC and Confusion Matrices when reporting on the performance of drift detectors \cite{HinderArteltHammer2020,HinderVaquetHammer2024}.

As shown in Section \ref{windowdilemma}, instance selection affects the approximated distributions that are used to detect concept drift. This means that any detection event raised by a drift detector is, at best, an indication that the partitioning of the selected instances exhibited what appears to be drift, not that concept drift has actually occurred. 

It is possible that the detected drift event corresponds to some real change in concept, but it is often infeasible to verify this. Despite being a classification task, we often do not know when real drift events happen in practice (i.e. we don't have drift labels) \cite{SouzaReisMaletzke2020}. This is supported by an over-reliance on synthetic data and the imputing of real-world datasets with known (i.e. labeled) drift. 

Because drift detection on \emph{actual} real-world datastreams is often an Unsupervised Learning task, performance improvement is used to imply that drift-based adaptive models perform better than non-adaptive models. This is a "Affirming the Consequent" fallacy. An observed performance improvement of a classifier equipped with a drift detector, relative to one without it, does not imply that the detector has actually identified concept drift. Performance improvement is neither necessary nor sufficient evidence of successful drift detection \cite{ReadZliobaite2025}.

Furthermore, it has been shown that concept drift detectors that detect more frequently, tend to perform better (up to a point). This is explicitly shown in Lukats et al.'s evaluation of unsupervised drift detectors \cite{LukatsZielinskiHahn2025} and it also present, although not explicitly stated or analyzed, in Souza et al.'s \cite{SouzaReisMaletzke2020} evaluation of performance-based detectors. It would appear that drift detectors that detect frequently are able to more frequently retrain classifiers on more recent data. Given the prominence of autocorrelation and time-dependency present in several data streams \cite{Zliobaite2013}, it is perhaps not surprising that frequent retraining results in better performing models. This has also been validated empirically with simple retraining regimes that do not make use of drift detectors \cite{Bifet2017}.
The effect of retraining a classifier depends on both the start and the end point of the training window. A frequent update of the end point allows timely incorporation of new observations. In contrast, updating the starting point triggers forgetting, which positively removes outdated/irrelevant instances at the risk of negatively excluding still relevant instances.

Another limiting aspect of drift detection that is not often discussed is what a "drift detection" means for the overall system. Liu et al. \cite{LuLiuDong2018} call this the "When" of drift detection, noting that few drift detectors go beyond that to determine "Where" and "How" the drift occurred. Webb et al. \cite{WebbHydeCao2016} provide several measures for quantifying the amount of drift a system experiences: 
(1) Drift Duration (how long is drift), 
(2) Drift Magnitude (how much drift),
(3) Path Length (how much total drift),
(4) Drift Rate (how frequently does drift occur)
but these measures themselves are susceptible to the Window Dilemma.

In terms of detectability, it may also be useful to think of:
(1) Persistence (duration for which a Concept remains stable) (may also be called Concept Stability \cite{WebbHydeCao2016}), 
(2) Sample Size (number of samples from a persistent concept),
(3) Sample Rate (the rate at which samples arrive from a persistent concept) (i.e. Sample Rate = Sample Size / Persistence).
From a theoretical perspective, a concept with a low sample rate may be fundamentally undetectable (this is known as blip drift \cite{WebbHydeCao2016}), and a concept with a high sample rate is easier to detect provided the appropriate windows are chosen. Practically, it is not clear how useful such ideas are without access to the ground truth drift labels.

Overall, our point is that simply "detecting drift" is insufficient, because (1) it is often impractical to verify whether these drift events happened, and (2) a drift event often tells very little about how the overall system or data stream itself is changing. Some drift may be irrelevant, some may be important, neither of which can be determined by viewing drift detection as binary (Yes/No) task, which we show empirically in the next Section.

\section{Experimental Design and Results} \label{exp}

In this Section we empirically evaluate the usefulness of drift-aware classifiers across a variety of benchmarking datasets. To achieve this, we will make use of two supervised drift detectors: (1) DDM \cite{GamaMedasCastillo2004} and (2) ADWIN \cite{BifetGavalda2007} and two unsupervised drift detectors: (1) D3 \cite{GozuaccikBuyukccakirBonab2019} and (2) IBDD \cite{SouzaParmezanChowdhury2021}. We chose ADWIN and DDM because of their popularity in literature, and D3 and IBDD because they were the best performing detectors in a recent benchmark evaluation \cite{LukatsZielinskiHahn2025}. For the D3 detector, we create two variants which use (1) Logistic Regression (D3-LR) and a (2) Hoeffding Tree (D3-HT) as the discriminatory classifier respectively. This was done because the choice of discriminator can have an impact of the performance of the detector.

For our base model, we make use of the popular (1) Naive Bayes (NB) and (2) Hoeffding Tree (HT) classifiers, which we also use as baselines. For each drift detector, we use each base model to create two adaptive-models per detector variant. For example, the DDM detector is paired with both the Naive Bayes (DDM-NB) and the Hoeffding Tree (DDM-HT) for a total of 10 detector-model pairs. When one of these detector-model pairs detects drift, the base model is reset and will start learning from scratch. We make also use of a periodic resetting technique as described by Bifet \cite{Bifet2017}. Every $N$ instances, the base model is simply reset. We do this for both base models (R-NB) and (R-HT). 

We extended our comparison to include also more sophisticated adaptive online learners. To that end, we also evaluate the Hoeffding Adaptive Tree \cite{BifetGalvalda2009} (HAT), Adaptive Random Forests \cite{GomesBifetRead2017} (ARF) and Aggregated Mondrian Forests \cite{MourtadaGaiffasScornet2021} (AMF). HAT and ARF have built-in mechanisms for utilizing concept drift detectors to adapt their internal structure (replacing subtrees in HAT, replacing base learners in ARF). AMF is a purely online technique that does not utilize concept drift detection.

We also include traditional batch learning techniques in the comparison with the aforementioned stream learning methods. We create three variants of the popular Random Forests \cite{Breiman2001} (RF): (1) a static Random Forest (S-RF) which is trained on the first $N=100$ instances of a stream and then never trained again, (2) a resetting Random Forest (R-RF) which is retrained every $N=50$ iterations on the past $N=50$ instances, and (3) an incremental Random Forest (I-RF) which is retrained every $N=50$ instances on all seen data. 

Lastly, we use of two baseline predictors: (1) Last Class (LC) and (2) Majority Class, which predict the last label seen and the majority class seen respectively. We include these baselines as it will reveal class imbalances and autocorrelation which are present in several data streams \cite{Zliobaite2011,Bifet2017}.

For evaluation, we use prequential evaluation (test-then-train) whereby a model will first make a prediction on an instance $X_i$ before it receives the label $Y_i$ to train on. We record the accuracy after each prediction and report the mean accuracy in this report. We acknowledge that accuracy is a potentially misleading metric in Drift Research \cite{Bifet2017, CerqueiraGomesBifet2023} and include the mean Cohen's Kappa statistic in Appendix \ref{sect:APP_KAPPA}. It does not change the overall message of our paper so we show accuracy for interpretability. 
We also use a psuedo-random number generator to initialize each model, ensuring reproducibility by providing the seeds.
The list of the parameter settings used are included in Appendix \ref{sect:APP_EXTENDED_DEETS}.

\subsection{Datasets}

Using a subset of the USP Data Stream Repository \cite{SouzaReisMaletzke2020}, we evaluate each model across 11 binary and multiclass data streams ranging over a variety of problem domains. These datasets are: Electricity (EL) \cite{Harries1999}, Forest Covertype (FC) \cite{BlackardDean1999}, Insects-Abrupt (balanced) (IA) \cite{SouzaReisMaletzke2020}, Insects-Incremental (balanced) (II) \cite{SouzaReisMaletzke2020}, Keystroke (KS) \cite{SouzaSilvaGamaBatista2015}, Luxembourg (LX) \cite{Zliobaite2011}, MIRS (MR) \cite{DauKeoghKamgar2018}, NOAA Weather (NW) \cite{DitzlerPolikar2012}, Ozone (OZ) \cite{DheeruTaniskidou2017}, Rialto (RT) \cite{LosingHammerWersing2016}, and Yoga (YG) \cite{DauKeoghKamgar2018}. 
Note that we do not make use of any purely synthetic datasets because we believe it diminishes from the message of the paper which is concerned about drift detection in the practical (real/semi-real) setting.
We include a short description of each dataset in Appendix \ref{sect:APP_EXTENDED_DEETS}.

\begin{table}[!p]
    \centering
    \begin{tabular}{|r|r|r|r|r|r|r|r|r|r|r|r|r|}
    \hline
        \multicolumn{1}{|c|}{} & \textbf{EL} & \textbf{FC} & \textbf{IA} & \textbf{II} & \textbf{KS} & \textbf{LX} & \textbf{MR} & \textbf{NW} & \textbf{OZ} & \textbf{RT} & \textbf{YG} & \textbf{MedRank} \\ \hline
        \multicolumn{13}{|c|}{} \\ 
        \multicolumn{13}{|c|}{Baselines} \\ \hline
        \multicolumn{1}{|r|}{\textbf{LC}} & 84,8 & \textbf{91,4} & 25,0 & 15,9 & 6,7 & 48,0 & 49,8 & 68,1 & 90,2 & 0,0 & 48,7 & 22 \\ \hline
        \multicolumn{1}{|r|}{\textbf{MC}} & 57,4 & 56,6 & 17,7 & 11,0 & 6,8 & 52,3 & 54,1 & 69,8 & 92,3 & 10,0 & 52,0 & 21 \\ \hline 
        \multicolumn{13}{|c|}{} \\ 
        \multicolumn{13}{|c|}{Naive Bayes} \\ \hline
        \multicolumn{1}{|r|}{\textbf{NB}} & 76,3 & 56,6 & 53,6 & 46,5 & 69,9 & 55,6 & 63,8 & 69,6 & 73,8 & 18,8 & 54,3 & 17 \\ \hline
        \multicolumn{1}{|r|}{\textbf{DDM-NB}} & 82,4 & 79,3 & 61,2 & 52,5 & 70,5 & 55,6 & 64,0 & 71,1 & 89,9 & 25,0 & 54,3 & 14 \\ \hline
        \multicolumn{1}{|r|}{\textbf{ADWIN-NB}} & 82,3 & 71,4 & 63,2 & 52,6 & 69,9 & 55,6 & 64,2 & 70,7 & 85,7 & 30,3 & 55,1 & 16 \\ \hline
        \multicolumn{1}{|r|}{\textbf{R-NB}} & 86,3 & 81,5 & 54,2 & 39,2 & 72,8 & 56,3 & 65,9 & 73,2 & 91,9 & 34,9 & 54,6 & 12 \\ \hline
        \multicolumn{1}{|r|}{\textbf{D3-LR-NB}} & 82,3 & 70,0 & 58,6 & 44,4 & 75,6 & 55,6 & 66,9 & 72,7 & 89,6 & 34,6 & 54,6 & 15 \\ \hline
        \multicolumn{1}{|r|}{\textbf{D3-HT-NB}} & 86,6 & 76,2 & 63,7 & 52,4 & 75,6 & 61,2 & 65,0 & 73,0 & 88,8 & 37,9 & 55,9 & 9 \\ \hline
        \multicolumn{1}{|r|}{\textbf{IBDD-NB}} & 86,7 & 82,0 & 59,8 & 46,4 & 72,5 & 56,6 & 66,1 & 71,0 & 88,1 & 23,6 & 56,8 & 10 \\ \hline
        \multicolumn{13}{|c|}{} \\ 
        \multicolumn{13}{|c|}{Hoeffding Trees} \\ \hline
        \multicolumn{1}{|r|}{\textbf{HT}} & 80,2 & 79,7 & 54,9 & 48,1 & 70,0 & 90,2 & 63,6 & 72,6 & 92,3 & 27,3 & 54,6 & 13 \\ \hline
        \multicolumn{1}{|r|}{\textbf{DDM-HT}} & 86,1 & 85,3 & 60,5 & 52,5 & 70,6 & 90,2 & 63,7 & 71,4 & 92,1 & 41,5 & 54,6 & 9 \\ \hline
        \multicolumn{1}{|r|}{\textbf{ADWIN-HT}} & 83,9 & 79,0 & 61,3 & 52,8 & 70,0 & 85,0 & 63,8 & 70,6 & 92,2 & 44,5 & 54,8 & 12 \\ \hline
        \multicolumn{1}{|r|}{ \textbf{R-HT}} & 86,0 & 87,3 & 40,0 & 33,3 & 69,7 & 64,0 & 63,0 & 72,6 & \textbf{92,8} & 53,9 & 52,5 & 13 \\ \hline
        \multicolumn{1}{|r|}{\textbf{D3-LR-HT}} & 84,3 & 80,4 & 46,7 & 41,0 & 77,5 & 90,2 & 65,3 & 71,8 & 92,4 & 51,0 & 53,2 & 10 \\ \hline
        \multicolumn{1}{|r|}{\textbf{D3-HT-HT}} & 86,4 & 85,7 & 59,4 & 51,8 & 74,9 & 87,5 & 64,5 & 72,2 & 92,4 & 55,8 & 55,4 & 9 \\ \hline
        \multicolumn{1}{|r|}{\textbf{IBDD-HT}} & 86,1 & 87,0 & 50,7 & 44,0 & 71,8 & 85,5 & 65,2 & 70,5 & 92,0 & 36,1 & 55,8 & 11 \\ \hline
        \multicolumn{1}{|r|}{\textbf{HAT}} & 84,6 & 79,7 & 57,7 & 49,4 & 70,5 & 91,9 & 63,7 & 73,8 & 92,3 & 29,3 & 55,1 & 11 \\ \hline
        \multicolumn{13}{|c|}{} \\ 
        \multicolumn{13}{|c|}{Online Random Forests} \\ \hline
        \multicolumn{1}{|r|}{\textbf{AMF}} & 85,0 & 90,5 & 67,2 & 57,5 & \textbf{88,7} & \textbf{95,4} & 73,7 & 76,4 & 92,0 & \textbf{82,8} & 75,5 & 3 \\ \hline
        \multicolumn{1}{|r|}{\textbf{ARF}} & \textbf{88,3} & 89,4 & \textbf{71,4} & \textbf{58,7} & 87,4 & 94,8 & 76,5 & 78,1 & 92,3 & 72,8 & 68,8 & 2 \\ \hline
        \multicolumn{13}{|c|}{} \\ 
        \multicolumn{13}{|c|}{Batch Random Forests} \\ \hline
        \multicolumn{1}{|r|}{\textbf{S-RF}} & 62,6 & 67,3 & 55,8 & 19,3 & 51,9 & 91,0 & 69,7 & 72,8 & 92,3 & 14,2 & 60,0 & 14 \\ \hline
        \multicolumn{1}{|r|}{\textbf{R-RF}} & 81,9 & 78,1 & 69,8 & 58,4 & 81,7 & 93,9 & 73,9 & 72,1 & 91,6 & 80,3 & 62,9 & 4 \\ \hline
        \multicolumn{1}{|r|}{\textbf{I-RF}} & 78,2 & 78,7 & 63,2 & 55,0 & 82,3 & 94,3 & \textbf{78,9} & \textbf{79,1} & 92,6 & 81,3 & \textbf{79,0} & 3 \\ \hline
    \end{tabular}%
    \caption{The Average Accuracy for each technique across all benchmark data steams investigated in this work. Bold values indicate the highest mean accuracy across the entire data stream. The median rank (MedRank, the lower the better) of each technique is also reported.}
    \label{tab:AvgAccValue}
\end{table}

\begin{table}[!p]
    \centering
    \begin{tabular}{|r|r|r|r|r|r|r|r|r|r|r|r|}
    \hline
        \multicolumn{1}{|c|}{} & \textbf{EL} & \textbf{FC} & \textbf{IA} & \textbf{II} & \textbf{KS} & \textbf{LX} & \textbf{MR} & \textbf{NW} & \textbf{OZ} & \textbf{RT} & \textbf{YG} \\ \hline
        \multicolumn{1}{|r|}{\textbf{AMF}} & 76,9 & 75,6 & 63,3 & 55,7 & 78,8 & 91,6 & 73,1 & 76,5 & 92,0 & 74,7 & 75,2 \\ \hline
        \multicolumn{1}{|r|}{\textbf{ARF}} & 81,1 & 69,7 & 66,7 & 56,8 & 77,0 & 93,7 & 74,1 & 77,2 & 92,2 & 63.3 & 67,8 \\ \hline
        \multicolumn{1}{|r|}{\textbf{R-RF}} & \textbf{81,9} & 78,1 & \textbf{69,8} & \textbf{58,4} & 81,7 & 93,9 & 73,9 & 72,1 & 91,6 & 80,3 & 62,9 \\ \hline
        \multicolumn{1}{|r|}{\textbf{I-RF}} & 78,2 & \textbf{78,7} & 63,2 & 55,0 & \textbf{82,3} & \textbf{94,3} & \textbf{78,9} & \textbf{79,1} & \textbf{92,6} & \textbf{81,3} & \textbf{79,0} \\ \hline
    \end{tabular}
    \caption{Experiments comparing the batch Reset and Incremental Random Forests (R-RF and I-RF) to the online Aggregated Mondrian Forest (AMF) and Adaptive Random Forest (ARF). These results show how AMF and ARF lose some of their efficacy when they follow the same label-delayed training regime as R-RF and I-RF.}
    \label{tab:SuppAccValue}
\end{table}

\subsection{Results}

Table \ref{tab:AvgAccValue} reports the results from our experiments. Note that while we report the median rank, the observed performance of a particular classifier is often data stream specific. For example, the ARF classifier is the most consistent classifier whereas the R-RF and I-RF classifiers perform exceptionally on some datasets, while performing comparatively poor on others (increasing their median rank). Nevertheless, we can still make several observations from these results.

On average, our results suggest that it is often better to have some form of retraining or resetting when a data stream is involved. This is identifiable by the poor performance of the non-adaptive Naive Bayes (NB), Hoeffding Tree (HT) and static Random Forest (S-RF).

In terms of the adaptive-classifiers, it is unclear which adaptive-technique is most appropriate. For the Naive Bayes, simple resetting (R-NB) works quite well for some datasets, but noticeably poor on both the Insect data streams (IA and II). For the Hoeffding Tree variants, resetting (R-HT) is less consistent, which may reflect in the HT's sensitivity to the data windows it is trained on.

Of the drift detectors, D3-HT performed the best for both the NB and HT base models. This is in line with the findings of Lukats et al. \cite{LukatsZielinskiHahn2025}. The HAT classifier performed worse than expected. Recall that HAT replaces poorly performing subtrees. This adaptive strategy does not seem to consistently outperform simply resetting the entire Hoeffding Tree when a drift event is detected. 

However, the classification strategy had the greatest impact on performance. More specifically, Adaptive Forest (tree-ensemble) techniques outperformed the single learner NB and HT variants. Interestingly, the drift-unaware Aggregated Mondrian Forest (AMF) performed competitively with the drift-aware Adaptive Random Forest (ARF). As noted before, the ARF classifier was the most consistent classifier we evaluated. We attribute this to ARF's ability to replace base learners that are no longer performing well. 

Interestingly, the resetting Random Forest (R-RF) and incremental Random Forest (I-RF) performed competitively on most of the datasets.\footnote{The Last Class classifier performed best on ForestCoverType because it abuses the known autocorrelation in the data stream. If we were to exclude this result, drift-ignorant classifiers would have performed the best on all but three of the data streams.}
This result is surprising because R-RF and I-RF are drift-unaware and for all intents and purposes, are batch learners applied in a stream learning context. It is often assumed that previous instances should be forgotten when drift is detected, our results suggest that keeping past data is often useful and periodically retraining a model on increasingly large training sets is often an effective strategy. There are some settings where this is not the case, but the performance of the R-RF classifier suggests that simple reset retraining on recent data is also viable.

Given that the R-RF and I-RF classifiers are at a disadvantage because they are only updated every $N$ instances whereas the stream learning classifiers could update themselves after every instance, we reran our experiments with the AMF and ARF models following the update regime used by R-RF and I-RF. The results reported in Table \ref{tab:SuppAccValue} are quite telling. Both AMF and ARF's consistency diminishes across many of the datasets. In fact, these results suggest that drift-aware stream learners are often inferior to simple batch learning procedures provided an appropriate classifier is chosen. Furthermore, the performance degradation of the AMF classifier suggests than periodically retraining your entire classifier from scratch has benefits over obeying the strict stream learning paradigm of learning from instances in the order they arrive. Overall, we believe these results reinforce our central claim that concept drift detection is broken. If simply keeping data to retrain on is the optimal strategy (as is the case for many of the data streams investigated), what is the purpose of a drift detector?

\section{Conclusions and Future Work}

This research highlighted two observations regarding Concept Drift Detection: 

(1) We showed how the process of selecting instances for drift detection innately determines if drift is perceived. This perceived drift may also be unreflective of a data stream at it's current point in time. The process of choosing the correct instance selection procedure is what we call the Window Dilemma. 

(2) Due to our inability to verify the accuracy to which our instance selection procedure approximates a concept at some time point, we argued that drift detection is ill-posed. We support this claim by synthesizing several well known limitations of drift detection such as our inability to verify all drift in practice.

We then empirically investigated a number of drift-aware adaptive classifiers (using various drift detectors) and compared them to simple online and drift-unaware adaptive classifiers across several data streams. Our results showed that the type of classifier used is often more important than drift-awareness.

We want to make it clear that we are not suggesting that any of theoretical discoveries surrounding concept drift are invalid, rather we wished to illustrate the apparent disconnect between the theory, and what we see in practice. We empirically showed that a state-of-the-art learning model (in combination with a little bit of domain knowledge) will often outperform a drift-aware model.
This complements traditional data stream research methods, which in the past mainly focused on efficient retraining under limited computational resources.
Our findings do not mean that drift detection is pointless, rather that its current purpose is misplaced. We believe our results are a call to action. It is not sufficient to reduce drift detection to a binary "When", and we should move towards developing drift understanding ("Where" and "How") \cite{LuLiuDong2018}, so that we may better exploit the data to create even better adaptive models. 
While the Window Dilemma itself is inherent to any choice of training window, insights from Ensemble Learning might offer some mitigation.
Further integration with areas such as Transfer Learning, Domain adaptation and Change Mining have potential \cite{KremplHoferWebbHullermeier2021}. In short, we believe learning how to exploit drift patterns (e.g. incremental) may be more useful than a simple drift detector could ever be.

\section*{Acknowledgments}
We are thankful for the financial support of the ICAI lab AI4Oversight. We thank Bernhard Pfahringer and members of the AI4Oversight Lab for early discussions, and the anonymous reviewers for their comments. 

\appendix

\section{Additional Experimental Details} \label{sect:APP_EXTENDED_DEETS}

All implementations were written in Python 3 with all Stream Learning techniques taken from the \verb|river=v0.23.0| \cite{Montiel2021} Python package and the batch learning Random Forests taken from \verb|SciKit-Learn=v1.7.2|. For the D3 detector, we used the river compatible implementation provided by the authors \cite{GozuaccikBuyukccakirBonab2019}. For the IBDD detector, we use the version provided by Lukats et al. \cite{LukatsZielinskiHahn2025} which we made \verb|river| compatible. 

We do not perform explicit parameter tuning for model on each data stream and instead use the default values provided by each in their respective libraries / implementations. We felt this was a more fair comparison than attempting to optimize each algorithm on each data stream. The default values are in general considered safe parameters to use in a variety of data stream scenarios.

Table \ref{tab:datasets} provides additional details regarding the datasets used in this work. Of importance are the Reset N and Retrain N columns. These refer to the reset rate of the periodic reset models (R-NB and R-HT) and the frequency at which the batch Random Forests (R-RF and I-RF) were retrained. For resetting, some datasets possess domain knowledge that we could exploit. For example, Luxembourg used \textit{N=32} which captures one month, Ozone uses \textit{N=84} which captures one month and NOAA we used \textit{N=120} which captures one season.

For retraining, it was merely a case of computational efficiency. The larger datasets are retrained less frequently.

The amount of samples in these datasets vary with the smallest, Keystroke, having 1,600 samples, and the largest data stream, Forest Covertype, having 581,012 samples.

\begin{table}[ht]
    \centering
    \begin{tabular}{|r|c|c|c|c|c|}
         \hline
         \textbf{Name} & \textbf{Classes} & \textbf{ Features} & \textbf{Samples} & \textbf{Reset N} & \textbf{Retrain N} \\
         \hline
         Electricity (EL) \cite{Harries1999} & 2 & 8 & 45,312 & 60 & 50  \\
         Forest Covertype (FC) \cite{BlackardDean1999} & 8 & 54 & 581,012 & 60 & 5000  \\
         INSECTS-Abrupt (balanced) (IA) \cite{SouzaReisMaletzke2020} & 6 & 33 & 52,848 & 60 & 500  \\
         INSECTS-Incremental (balanced) (II) \cite{SouzaReisMaletzke2020} & 6 & 33 & 57,018 & 60 & 500  \\
         Keystroke (KS) \cite{SouzaSilvaGamaBatista2015} & 4 & 10 & 1,600 & 60 & 50   \\
         Luxembourg (LX) \cite{Zliobaite2011} & 2 & 30 & 1,901 & 32 & 50   \\
         MIRS (MR) \cite{DauKeoghKamgar2018} & 2 & 3,600 & 4,260 & 60 & 50  \\
         NOAA Weather (NW) \cite{DitzlerPolikar2012} & 2 & 8 & 18,159 & 120 & 50  \\
         Ozone (OZ) \cite{DheeruTaniskidou2017}  & 2 & 72 & 2,534 & 84 & 50  \\
         Rialto (RT) \cite{LosingHammerWersing2016} & 10 & 27 & 82,250 & 60 & 50  \\
         Yoga (YG) \cite{DauKeoghKamgar2018} & 2 & 426 & 3,300 & 60 & 50\\
         \hline
    \end{tabular}
    \caption{Summary of the Datasets used in this paper.}
    \label{tab:datasets}
\end{table}

\begin{table}[ht]
    \centering
    \begin{tabular}{|r|r|r|r|r|r|r|r|r|r|r|r|r|}
    \hline
        \multicolumn{1}{|c|}{} & \textbf{EL} & \textbf{FC} & \textbf{IA} & \textbf{II} & \textbf{KS} & \textbf{LX} & \textbf{MR} & \textbf{NW} & \textbf{OZ} & \textbf{RT} & \textbf{YG} & \textbf{MedRank} \\ \hline
        \multicolumn{13}{|c|}{} \\
        \multicolumn{13}{|c|}{Baselines} \\ \hline
        \multicolumn{1}{|r|}{\textbf{LC}} & 68,9 & \textbf{85,8} & 9,4 & -1,0 & -24,3 & -4,4 & -1,3 & 24,1 & \textbf{29,2} & -11,1 & -3,1 & 21 \\ \hline
        \multicolumn{1}{|r|}{\textbf{MC}} & 0,0 & 6,9 & 0,2 & -6,8 & -23,9 & -0,6 & 0,0 & 0,0 & 0,0 & 0,0 & -1,4 & 21 \\ \hline
        \multicolumn{13}{|c|}{} \\
        \multicolumn{13}{|c|}{Naive Bayes} \\ \hline
        \multicolumn{1}{|r|}{\textbf{NB}} & 49,2 & 0,4 & 44,4 & 35,8 & 60,0 & 6,8 & 22,7 & 32,6 & 19,5 & 9,8 & 9,2 & 17 \\ \hline
        \multicolumn{1}{|r|}{\textbf{DDM-NB}} & 63,5 & 62,9 & 53,4 & 43,0 & 60,8 & 6,8 & 23,2 & 35,9 & 22,4 & 16,7 & 9,2 & 14 \\ \hline
        \multicolumn{1}{|r|}{\textbf{ADWIN-NB}} & 63,4 & 48,1 & 55,9 & 43,1 & 60,0 & 6,8 & 23,8 & 37,0 & 25,0 & 22,5 & 11,1 & 14 \\ \hline
        \multicolumn{1}{|r|}{\textbf{R-NB}} & 72,0 & 68,0 & 44,5 & 27,0 & 63,8 & 10,5 & 30,1 & 35,7 & 21,4 & 27,6 & 8,8 & 9 \\ \hline
        \multicolumn{1}{|r|}{\textbf{D3-LR-NB}} & 63,3 & 44,8 & 50,0 & 33,3 & 67,6 & 6,8 & 31,0 & 36,2 & 24,8 & 27,4 & 9,4 & 11 \\ \hline
        \multicolumn{1}{|r|}{\textbf{D3-HT-NB}} & 72,6 & 58,2 & 56,3 & 42,9 & 67,6 & 18,9 & 25,7 & 36,5 & 24,1 & 30,9 & 12,5 & 7 \\ \hline
        \multicolumn{1}{|r|}{\textbf{IBDD-NB}} & 72,7 & 68,8 & 51,5 & 35,6 & 63,5 & 9,0 & 29,3 & 29,2 & 22,3 & 15,2 & 13,9 & 10 \\ \hline
        \multicolumn{13}{|c|}{} \\
        \multicolumn{13}{|c|}{Hoeffding Trees} \\ \hline
        \multicolumn{1}{|r|}{\textbf{HT}} & 58,6 & 64,2 & 45,7 & 37,7 & 60,1 & 80,3 & 22,1 & 30,8 & 0,0 & 19,3 & 9,1 & 15 \\ \hline
        \multicolumn{1}{|r|}{\textbf{DDM-HT}} & 71,5 & 74,6 & 52,5 & 43,0 & 60,8 & 80,3 & 22,5 & 31,7 & 8,0 & 35,0 & 9,1 & 9 \\ \hline
        \multicolumn{1}{|r|}{\textbf{ADWIN-HT}} & 66,8 & 63,5 & 53,5 & 43,3 & 60,1 & 69,7 & 22,9 & 28,3 & 3,4 & 38,3 & 9,7 & 13 \\ \hline
        \multicolumn{1}{|r|}{\textbf{R-HT}} & 71,4 & 78,3 & 27,7 & 20,0 & 59,6 & 27,2 & 23,8 & 26,5 & 18,9 & 48,8 & 4,0 & 13 \\ \hline
        \multicolumn{1}{|r|}{\textbf{DR-LR-HT}} & 67,5 & 65,8 & 35,7 & 29,2 & 70,0 & 80,3 & 27,6 & 26,8 & 19,1 & 45,5 & 6,3 & 10 \\ \hline
        \multicolumn{1}{|r|}{\textbf{D3-HT-HT}} & 72,0 & 75,3 & 51,1 & 42,2 & 66,6 & 74,8 & 24,6 & 27,6 & 19,1 & 50,9 & 11,1 & 9 \\ \hline
        \multicolumn{1}{|r|}{\textbf{IBDD-HT}} & 71,6 & 77,8 & 40,6 & 32,8 & 62,4 & 70,8 & 27,2 & 23,3 & 10,8 & 29,0 & 12,0 & 11 \\ \hline
        \multicolumn{1}{|r|}{\textbf{HAT}} & 68,4 & 64,5 & 49,2 & 39,3 & 60,6 & 83,2 & 23,0 & 36,3 & 0,0 & 21,4 & 10,2 & 11 \\ \hline
        \multicolumn{13}{|c|}{} \\
        \multicolumn{13}{|c|}{Adaptive Random Forests} \\ \hline
        \multicolumn{1}{|r|}{\textbf{AMF}} & 68,9 & 83,1 & 60,3 & 49,0 & \textbf{85,0} & \textbf{90,8} & 46,2 & 36,7 & 14,3 & \textbf{80,9} & 50,6 & 3 \\ \hline
        \multicolumn{1}{|r|}{\textbf{ARF}} & \textbf{75,9} & 81,3 & \textbf{65,4} & \textbf{50,5} & 83,4 & 89,4 & 51,8 & 42,2 & 3,9 & 69,8 & 37,4 & 2 \\ \hline
        \multicolumn{13}{|c|}{} \\
        \multicolumn{13}{|c|}{Batch Random Forests} \\ \hline
        \multicolumn{1}{|r|}{\textbf{S-RF}} & 28,9 & 43,4 & 46,5 & 3,2 & 36,1 & 81,2 & 39,9 & 18,9 & 0,0 & 4,7 & 20,9 & 20 \\ \hline
        \multicolumn{1}{|r|}{\textbf{R-RF}} & 62,9 & 60,2 & 63,3 & 50.0 & 75,9 & 87,4 & 47,1 & 29,2 & 8,1 & 78,1 & 25,4 & 4 \\ \hline
        \multicolumn{1}{|r|}{\textbf{I-RF}} & 54,6 & 61,8 & 55,5 & 45,9 & 76,6 & 88,2 & \textbf{57,5} & \textbf{45,6} & 11,1 & 80,3 & \textbf{57,9} & 3 \\ \hline
    \end{tabular}
    \caption{The Average Cohen's Kappa statistic for each technique across all benchmark data steams. Bold values indicate the highest mean kappa score across the entire data stream. The median rank (MedRank) of each technique is also reported. Note that we have scaled the kappa statistic by 100 for readability. For example, a kappa score of 0,543 would be reported as 54,3.}
    \label{tab:avgKappvalue}
\end{table}

\section{Average Kappa Scores} \label{sect:APP_KAPPA}

As mentioned earlier, Accuracy can often be misleading when interpreting the results of classifiers on data streams with drift \cite{Bifet2017, CerqueiraGomesBifet2023}. Table \ref{tab:avgKappvalue} reports these results. Overall, the main message of the paper remains constant and we include these results for completeness.

\section{Cold vs. Warm Starts} \label{sect:APP_COLD_START}

\begin{table}[ht]
    \centering
    \begin{tabular}{|r|r|r|r|r|r|r|r|r|r|r|r|r|}
    \hline
        \multicolumn{1}{|c|}{} & \textbf{EL} & \textbf{FC} & \textbf{IA} & \textbf{II} & \textbf{KS} & \textbf{LX} & \textbf{MR} & \textbf{NW} & \textbf{OZ} & \textbf{RT} & \textbf{YG} & \textbf{MedRank} \\ \hline

        \multicolumn{13}{|c|}{} \\ 
        \multicolumn{13}{|c|}{Batch Random Forests (Warm Start)} \\ \hline
        \multicolumn{1}{|r|}{\textbf{S-RF}} & 62,5 & 68,6 & 57,5 & 20,2 & 64,6 & 94,7 & 71,3 & 72,5 & 86,7 & 14.3 & 61,0 & 20 \\ \hline
        \multicolumn{1}{|r|}{\textbf{R-RF}} & 82,1 & 79,5 & \textbf{71,8} & \textbf{60,1} & 91,0 & 96,9 & 74,7 & 71,9 & 89,2 & 80,6 & 63,5 & 3 \\ \hline
        \multicolumn{1}{|r|}{\textbf{I-RF}} & 78,4 & 80,2 & 65,1 & 56,7 & \textbf{91,6} & \textbf{97,4} & \textbf{79,9} & \textbf{79,0} & 90,1 & 83,4 & \textbf{80,1} & 1 \\ \hline
        \multicolumn{13}{|c|}{} \\ 
        \multicolumn{13}{|c|}{Batch Random Forests (Cold Start)} \\ \hline
        \multicolumn{1}{|r|}{\textbf{S-RF}} & 62,6 & 67,3 & 55,8 & 19,3 & 51,9 & 91,0 & 69,7 & 72,8 & 92,3 & 14,2 & 60,0 & 14 \\ \hline
        \multicolumn{1}{|r|}{\textbf{R-RF}} & 81,9 & 78,1 & 69,8 & 58,4 & 81,7 & 93,9 & 73,9 & 72,1 & 91,6 & 80.3 & 62,9 & 4 \\ \hline
        \multicolumn{1}{|r|}{\textbf{I-RF}} & 78,2 & 78,7 & 63,2 & 55,0 & 82,3 & 94,3 & \textbf{78,9} & \textbf{79,1} & 92,6 & 81,3 & \textbf{79,0} & 3 \\ \hline
    \end{tabular}%
    \caption{The Average Accuracy for each batch learning technique across all benchmark data steams. This Table compares the effect that a cold (using a majority class classifier) vs. warm start approach has on the overall performance of the batch learners. The median rank (MedRank) is calculated using all the results from our initial evaluation in Table \ref{tab:AvgAccValue}. Bold values indicate results that would've achieved top results across those same initial evaluations. For example, the warm start R-RF model achieves top performance on the IA and II data sets, but the cold start approach does not. }
    \label{tab:ColdStart}
\end{table}

Another set of experiments we ran removed the cold start from the batch learning methods (i.e. the use of a Majority Class classifier on the first set of N instances). Table \ref{tab:ColdStart} shows the difference in results. Interestingly, the average accuracy for some data sets increases when the Majority Class classifier is used. This is largely found on the highly imbalanced or auto-correlated data streams where a majority class classification scheme is beneficial. In most cases, the performance of the warm starts is better. This makes sense as making fewer predictions and starting with some information about the data stream (the warm start) is better than starting with none (the cold start).

The implications of these results are interesting. On the one hand, it is reasonable to assume that some learning tasks would provide you with some initial data to train on (warm start), but it does remove the added context regarding an algorithm's sample efficiency (how many instances does it take to learn the initial concept) that the cold start provides. Overall, both methods have their merits, but should be clearly stated, so that the limitations of such evaluation decisions are transparent.

\bibliographystyle{unsrt}  
\bibliography{bib}  

\end{document}